\documentclass{IEEEcsmag}

\usepackage[colorlinks,urlcolor=blue,linkcolor=blue,citecolor=blue]{hyperref}
\usepackage{booktabs}
\usepackage{upmath}
\usepackage{multirow}

\begin{document}


\title{Long-Short Ensemble Network for Bipolar Manic-Euthymic State Recognition Based on Wrist-worn Sensors}

\author{Ulysse Côté-Allard\,$^{1}$, Petter Jakobsen\,$^{2, 3}$, Andrea Stautland\,$^{3}$, Tine Nordgreen\,$^{4, 5}$, Ole Bernt Fasmer\,$^{2, 3},$ Ketil Joachim Oedegaard\,$^{2, 3}$, Jim Torresen\,$^{1}$}
\affil{$^{1}$ Department of Informatics, University of Oslo, Oslo, Norway\\
 $^{2}$ NORMENT, Division of Psychiatry, Haukeland University Hospital, Bergen, Norway\\
$^{3}$ Department of Clinical Medicine, University of Bergen, Norway\\
$^{4}$ Department of Clinical Psychology, Faculty of Psychology, University of Bergen, Bergen, Norway\\
$^{5}$ Division of Psychiatry, Haukeland University Hospital, Bergen, Norway\\}


\begin{abstract}
Manic episodes of bipolar disorder can lead to uncritical behaviour and delusional psychosis, often with destructive consequences for those affected and their surroundings. Early detection and intervention of a manic episode are crucial to prevent escalation, hospital admission and premature death. However, people with bipolar disorder may not recognize that they are experiencing a manic episode and symptoms such as euphoria and increased productivity can also deter affected individuals from seeking help. This work proposes to perform user-independent, automatic mood-state detection based on actigraphy and electrodermal activity acquired from a wrist-worn device during mania and after recovery (euthymia). This paper proposes a new deep learning-based ensemble method leveraging long (20h) and short (5 minutes) time-intervals to discriminate between the mood-states. When tested on 47 bipolar patients, the proposed classification scheme achieves an average accuracy of 91.59\% in euthymic/manic mood-state recognition.

\end{abstract}

\maketitle


\chapterinitial{Bipolar Disorder} is a severe mental disorder characterized by intense periodic mood fluctuations, lifelong disability and a high disease burden which affects more than 1\% of the global population~\cite{overview_bipolar_disorder, carvalho2020bipolar}.

People with bipolar disorder have a mortality risk twice as high as the general population, due to somatic comorbidities and suicide rates 20 to 30 times higher than the general population~\cite{carvalho2020bipolar}.
Bipolar disorder is usually divided into two subgroups, Bipolar I and Bipolar II. Bipolar I is defined by the presence of manic episodes, typically characterized by increased energy, inflated self-esteem, increased need to pursue goal-directed actions, reduced subjective need for sleep, and often associated with the presence of hallucinations and delusions. The elevated mood defining Bipolar II is hypomania, a less severe form of mania, and without hallucinations and delusions. Another difference is that at least one major depressive episode is needed for the diagnosis of Bipolar II but not Bipolar I. The presence of depressive episodes, which are typically characterized by diminished initiative and energy, as well as disturbed sleep patterns are nevertheless common in Bipolar I. The neutral state \textit{euthymia} can be characterized as the in-between state that neither meets the criteria for depression nor mania/hypomania.

Early help and intervention is an important factor in mitigating the risks associated with mania~\cite{bauer2018areas}. However, it can be hard for the affected person to realise that they are experiencing an episode~\cite{overview_bipolar_disorder}. Further, even when recognizing that a manic episode is occurring, the sense of euphoria and increased productivity can be dissuading factors in seeking help.

As no biomarker has yet been approved for the diagnosis of bipolar disorder~\cite{overview_bipolar_disorder}, current practices in assessing mood episodes focus on subjective observation in conjunction with semi-structured clinical rating scales~\cite{petter_actigraphy_ml_depression_classification}. Consequently, it remains challenging to perform efficient targeted interventions, due to the delicate balance between adequately monitoring the patient and moderating the impact of repeated appointments on the healthcare system and the patient's life. Changes in mood triggered by an affective disorder are not only associated with changes in behaviour, but are also reflected in several biological processes, such as in the autonomic nervous system~\cite{overview_bipolar_disorder, petter_actigraphy_ml_depression_classification, bassett_literature_review_hrv_in_bipolar}.
As a result, much effort has been deployed in characterizing mood-states in affective disorders from various biosignals (e.g. electrodermal activity (EDA), actigraphy, electrocardiogram),
with the aim of automatically identifying state-change without human intervention. To achieve this goal however, the system used to record the biosignals must be non-intrusive to allow continuous recording without affecting the patient's daily life. Smartwatches and smart wristbands are especially well suited for such an application as in addition to being non-intrusive, they are easy to setup, commonly available and relatively inexpensive.
Consequently, this work focuses on the problem of manic-euthymic automatic state recognition using biosignals recorded from a wrist-worn wearable. For state-recognition, the sensors considered are a 3-axis accelerometer (actigraphy), EDA and a photoplethysmography (PPG) (from which the heart rate (HR) can be derived). Further, this work considers the setting where no data for training is available from the patient that is to be predicted on. This is necessary for the system to be calibration-free and user-independent. 

The literature on state recognition in affective disorders primarily focuses on feature engineering, with the goal of characterizing a segment generated from a given modality (e.g. heart rate, actigraphy, speech) in a discriminative way. While these types of approaches have been shown to be able to discriminate between different states~\cite{petter_actigraphy_ml_depression_classification},
they often do not explicitly consider the temporality of the characterized segment. Contrastively, Time Series Classification (TSC) algorithms are made specifically to leverage this temporal information.
For multivariate TSC, InceptionTime is a method based on convolutional networks which was shown to achieve state of the art results for real-time multivariate TSC applications~\cite{time_series_inception_2020}. 
As such, one of this work's contributions is to divide a multimodal segment into multiple sub-segments, from which meaningful features are extracted before applying an InceptionTime-based architecture to perform automatic manic-euthymic state recognition for never-seen-before patients. 

The type of information derived from the characterization of biosignals is dependent on the considered timespan (e.g. seconds, minutes, hours)~\cite{petter_actigraphy_ml_depression_classification, shaffer_hrv_features}. Consequently, another contribution of this work is to employ an ensemble of networks which are fed features extracted from both minute-long and hour-long intervals to leverage the information extracted from both horizon lengths. Note that in this work, the term \textit{network} is used in a machine learning context to refer to deep learning models.

\section{Data Acquisition and Preprocessing}
\label{dataset_section}
\subsection{Participants and Data Acquisition}
As a first step in the goal of automatically detecting manic episodes, this work focuses on a dataset that was recorded 
in a two-phased clinical study of bipolar disorder. All participants were diagnosed according to the International Classification of Diseases (ICD)-10~\cite{overview_bipolar_disorder}. 
58 participants were included and of these 28 were recorded when hospitalized due to an ongoing manic episode (ICD-10 diagnosis F31.1 (current episode manic without psychotic symptoms) and F31.2 (current episode manic with psychotic symptoms)). The clinical psychiatrists residing at the two closed affective wards at Haukeland University Hospital
suggested potential candidates after assessing their ability to consent. 

A group of 30 euthymic patients, not overlapping with the manic group, were also included in this study. These participants were additional participants from the first part of the study (when discharged from the hospital), or recruited from the hospitals’ outpatient clinic or the local advocacy group for bipolar disorder.






Inclusion criteria for both phases of the study were Norwegian speaking individuals between 18 and 70 years diagnosed with bipolar disorder, able to comply with instructions and having an IQ above 70. Exclusion criteria were previous head trauma needing hospital treatment, having an organic brain disorder, substance dependence (excluding nicotine), or being in a withdrawal state. The study protocol was approved by The Norwegian Regional Medical Research Ethics Committee West (2017/937). Written informed consent was obtained from all participants and no financial compensation/treatment perks were provided. All patients (except two in the euthymic group) were taking various combinations of prescribed medications.

The patients' mood-states were established at inclusion and at regularly repeated clinical assessments using the Young Mania Rating Scale (YMRS)~\cite{overview_bipolar_disorder}.
YMRS rates the severity of mania based on clinical observations and the patients’ subjective description of their state. The total score spans from 0 to 60, and a YMRS score below 10 is considered as being in remission, or in an euthymic state~\cite{malhi2007neuropsychological}. The participants were also assessed with the Montgomery Asberg Depression Rating Scale (MADRS)~\cite{overview_bipolar_disorder}, a commonly used scale for measuring the presence and severity of an ongoing depression. MADRS scores span between 0 and 60, and scores below 10 are defined as the absence of depression~\cite{hawley2002defining}.
 For the euthymic participants, the bipolar diagnosis was validated using the Mini-International Neuropsychiatric Interview (MINI) version 6.0.0~\cite{sheehan1998mini}. 


Table~\ref{table:demographic_characteristic}
presents the demographic characteristics for both groups. 


\begin{table*}[!htbp]
\caption{Characteristics and demographics of the two analyzed patient groups}
\centering
\begin{tabular}{ccccc}
\hline
 &  & Manic & Euthymic & p \\ \hline
\textbf{Group Size, Age and Sex} &  &  &  &  \\
 & N & 28 & 30 &  \\
 & Mean Age (SD) & 44 (15) & 43 (14) & NS* \\
 & Age Range (minimum - maximum) & 18-70 & 23-67 &  \\
 & Sex (Percent Women) & 57\% & 67\% & NS** \\ \hline
\textbf{Marital Status} &  &  &  & NS** \\
 & Single/Divorced & 54\% & 53\% &  \\
 & Married/Cohabiting & 46\% & 47\% &  \\ \hline
\textbf{Employment Status} &  &  &  & NS** \\
 & Employed/Student & 43\% & 63\% &  \\
 & Unemployed & 11\% & 17\% &  \\
 & Disability Benefit/Retired & 46\% & 20\% &  \\ \hline
\textbf{Level of Education} &  &  &  & NS** \\
 & Junior High School & 32\% & 20\% &  \\
 & High School/Vocational Studies & 25\% & 37\% &  \\
 & University/Higher Education & 36\% & 43\% &  \\
 & Unknown & 7\% & 0\% &  \\ \hline
\textbf{Diagnostic Status} &  &  &  &  \\
 & Diagnoses (BP1/BP2) & 28/-$^a$ & 17/13 & \textless{}0.001** \\
 & Mean Score YMRS (SD) & 24.1 (3.7) & 3.2 (2.1) & \textless{}0.001* \\
 & Mean Score MADRS (SD) & 6.1 (4.4) & 4.1 (3.5) & NS* \\
 & Season for E4 Recording (percent summer)$^b$ & 43 & 50 & NS** \\ \hline
\textbf{Psychopharmacological Treatment} &  &  &  & 0.015** \\
 & Mood Stabilizers & 86\% & 80\% &  \\
 & Antipsychotics & 96\% & 47\% &  \\
 & Antidepressant & 7\% & 23\% &  \\
 & Benzodiazepines & 29\% & 10\% &  \\ \hline
\end{tabular}
\\[4pt]
Abbreviations: SD = standard deviation, NS = Not Significant, BP1 = Bipolar disorder type 1, BP2 = Bipolar disorder type 2, YMRS = Young Mania Rating Scale, MADRS = Montgomery Asberg Depression Rating Scale, - =Not Applicable.\\
* Independent Samples t-test with Levene’s test for Equality of Variance. Null hypothesis rejected at p$<$0.05.  \\
** Pearson's chi-squared test. Null hypothesis rejected at p$<$0.05.\\
$^a$ Clinical ICD-10 diagnosis given at hospitalization for the current manic episode, either: F31.1, current episode manic without psychotic symptoms (39\%) or F31.2, current episode manic with psychotic symptoms (61\%).\\
$^b$ Summer defined as the half-year period between the vernal and autumnal equinoxes.
\label{table:demographic_characteristic}
\end{table*}

The data used in this work was recorded with the Empatica E4 wristband worn on the dominant wrist for 24h. The device provides a 3-axis accelerometer, an EDA sensor, a skin-temperature sensor and a PPG.
\subsection{Post-recording exclusion}
This study aimed to limit the impact of the recording process on the participants' behavior. Therefore, besides being asked to wear the smart wristband, participants continued their treatment unhindered by the research protocol. Consequently, depending on when the next day assessment took place, the total recording period varied between participants and could span less than 24h. Additionally, some participants removed their wristband during recording, sometimes multiple times and for multiple hours. Therefore, manual segmentation based on skin-temperature and accelerometer was performed to identify and remove the data recorded when the wristband was off.
Because of these two factors aggregating, three participants (all manic) did not reach the minimum amount of data defined within this study ($>$20h) and were not considered when reporting results. 

\subsection{Dataset Segmentation}
Acquiring data in a clinical context is a laborious process, often making the creation of large datasets impractical. Further, as the samples are not independent and identically distributed (i.i.d.), special care has to be taken to avoid data leakage (i.e. information contained within the test set indirectly being used during training). Consequently, within this work, data is compartmentalized such that samples from the same individual will only be considered within the same set (i.e. train/validation/test set). Further, a subset of the recorded dataset was reserved for data exploration, architecture building and hyperparameter optimization. This subset, dubbed the \textit{exploration dataset} is comprised of three manic and five euthymic randomly selected participants. An additional two manic participants  come from two of the three previously excluded individuals (as $>$18h of recording was available for both). This was done to minimize the amount of participants that had to be taken out and to leverage otherwise discarded data. 


The dataset containing the remaining 47 participants (22 manic and 25 euthymic) will be referred to as the \textit{main dataset}. Due to the limited amount of participants
, leave-one-out cross-validation is employed for evaluating the different methods considered.
In other words, to evaluate a classifier, 47 independent rounds of training will be performed where the held-out test set will correspond to a different individual each time. Further, the exploration dataset is concatenated with the main dataset's training set to increase the amount of training data which can facilitate better generalization.
Due to the stochastic nature of the considered algorithms, all results are reported as an average of 20 runs.


\section{Data Processing}
\label{feature_extraction}

The following section details the data processing employed for each modality and presents the different feature sets considered. Note that skin-temperature can be influenced by external factors (e.g. ambient temperature), which can lead to data leakage (e.g. higher room temperature on average for a given group). 
As this factor was not controlled for, skin-temperature's contribution in distinguishing the mood-state is not investigated.

\subsection{Processing of the different modalities}
Data processing of the biosignal was facilitated by the NeuroKit2~\cite{Makowski2021neurokit} library in Python. 

\subsubsection{Electrodermal Activity}
The EDA employed in the wristband has a sampling frequency of 4Hz and a range between 0.01 and 100 $\mu$Siemens.

During processing, a low-pass butterworth filter of order 4 at 1.5Hz is applied to better capture both the tonic and phasic component of the signal~\cite{eda_frequency_band_between_0_05_and_1_5}. 
From the cleaned signal, a high-pass butterworth filter of order 2 at 0.05Hz is applied to extract the phasic component of the signal~\cite{Makowski2021neurokit, eda_frequency_band_between_0_05_and_1_5}. Skin Conductance Response (SCR) peaks are then identified by extracting the local maxima of the filtered signal, rejecting peaks with an amplitude below 10\% of the standard deviation from the mean of the amplitude as implemented in~\cite{Makowski2021neurokit}. 



\subsubsection{Photoplethysmograph and Heart Rate}
The wristband's PPG employs a green and a red light-emitting diode (LED). The E4 uses a black box algorithm to fuse the information retrieved from the green and red exposure to limit the impact of motion artefacts.
The black box algorithm's output is what is made available at a sampling rate of 64Hz. Within this work, a band-pass butterworth filter of order 3 was applied between 0.5 and 8Hz to the signal. The systolic peaks were then extracted from the filtered signal based on the method described in~\cite{ppg_systolic_peaks_extraction} and implemented in~\cite{Makowski2021neurokit}. The distances between these peaks are referred to as NN to emphasize the fact that abnormal beats have been removed~\cite{shaffer_hrv_features}. 

The HR is also made available by the E4 at a sampling rate of 1Hz and represented the average HR values computed in a span of 10 seconds.

\subsubsection{Actigraphy}
The 3-axis accelerometer integrated in the E4 has a range of $\pm$2g and is cadenced at 32Hz. For each participant, each data point was processed as follows: 
\begin{equation}
 \sqrt{x^2 + y^2 + z^2}-1g
\end{equation}
 Where x, y and z represent the recorded value for their associated axis and $1g$ represents the gravitational constant.

\section{Feature Extraction}
\subsection{Electrodermal Activity Feature Set}
Two features were extracted from the EDA modality. First, the autocorrelation with a lag of 4 was computed from the filtered low-pass EDA signal as suggested in~\cite{EDA_SRC_Peaks_extraction_vanHalem}. The second feature was extracted by taking the mean amplitude of the SCR peaks.
\subsection{Heart Rate Variability Feature Set}
The sample entropy (SampEn) was extracted to measure the level of predictability in successive NN intervals~\cite{shaffer_hrv_features}. The standard deviation of the NN intervals (SDNN)~\cite{shaffer_hrv_features} was also calculated. Note that popular features such as the root mean square of successive differences (RMSSD)~\cite{shaffer_hrv_features} and Low-Frequency/High-Frequency ratio~\cite{shaffer_hrv_features} were not considered as they are particularly noisy when computed from a PPG signal~\cite{scientific_reports_hrv_ppg_error}. Consequently, the feature set extracted for the Heart Rate Variability (HRV) is as follows: 


\begin{equation}
\left[SDNN, SampEn\right]
\end{equation}

\subsection{Actigraphy and Heart Rate Feature Sets}
Multiple feature sets were considered for the characterization of both the processed actigraphy and heart rate.

\subsubsection{Bipolar Complexity Variability Features Set}
The Bipolar Complexity-Variability (BCV) feature set is derived from~\cite{Petter_complexity_variability_2021} and is defined as follows: 
\begin{equation}
\left[\frac{\sigma}{\mu}, \frac{RMSSD}{SD}, SampEn\right]
\end{equation}

Where $\mu$ and $\sigma$ correspond to the mean and standard deviation of the signal, while RMSSD corresponds to the root mean square of successive difference.


\subsubsection{TSD}
The initial features proposed in~\cite{TSD} are considered as a features set and referred to as Temporal-Spatial Descriptors (TSD). TSD consists of: the Root squared zero, second and fourth moments as well as the Sparseness, Irregularity Factor, Coefficient of Variation and the Teager-Kaiser energy operator. 

In addition, a new feature set proposed in this work is the combination of TSD with BCV, which will be referred to as the TSD-BCV feature set. 

\section{Mood-State Classification Methods}
Two types of intervals from which to compute the different feature sets are considered: long (20h) and short (5 minutes). 
The following subsection provides a thorough description of the classifiers used for both intervals and their combination. 

\subsection{Long Interval Classification}
Sequences lasting 20h were selected in this work as a balancing act between including as many of the participants as possible for evaluation (as their recording needed to be at least that long) and being as close to a full day cycle as possible. The previously presented feature sets are thus computed directly from these long intervals for each participant. When considering multiple modalities simultaneously, features from each sensor are concatenated together into a single vector. As a form of data augmentation, a sliding window with an overlap of 19.5h is applied to generate the examples from each participant. This data augmentation procedure resulted in an average of $\sim$13 examples per participant.

For each fold in the leave-one-out cross-validation scheme, each feature is scaled between -1 and 1 using min-max scaling. Note that the minimum and maximum values are obtained from the training set and the min-max normalization is performed on both the training and test set. The following eight classifiers are then considered for mood-state classification: K-Nearest Neighbors (KNN), Linear Discriminant Analysis (LDA), Quadratic Discriminant Analysis (QDA), Decision Tree (DT), Random Forest (RF), AdaBoost and Support Vector Machine (SVM) both with a Linear and Radial Basis Function (RBF) kernel. Class weights were balanced to account for under/overrepresentation of a given class.
Hyperparameter selection is performed using random search
with 50 candidates.
The validation set employed for the random search is extracted from the current training set fold by randomly selecting 2 manic and 2 euthymic participants. The hyperparameters considered for each classifier are presented in Appendix-A. The classifiers' implementation comes from scikit-learn (v0.24.1) in Python~\cite{scikit-learn}. 

\subsection{Short Interval Classification}
Instead of characterizing the signal by extracting features over long intervals, this classification approach proposes considering much shorter intervals (five minutes) as subwindows of the full example from which to extract the features. For each fold in the leave-one-out cross-validation scheme, each feature is then scaled between -1 and 1 using min-max scaling, as previously described. An example is then created by aggregating consecutive subwindows to form a $F_T\times W$ matrix. Where $F_T$ represents the number of input features
and $W$ being the number of subwindows forming the example. The idea is then to perform feature learning via an InceptionTime Network to discriminate between the different mood-states. Note that due to the structure of the network's architecture employed, it is possible to train with examples of varying lengths (i.e. number of subwindows). As such, the examples created vary in length between 20h and 24h using increments of 40 minutes. Additionally, examples were created with a sliding window using increments of 25 minutes. This data augmentation procedure yields an average training set containing $\sim$4000 examples. As from the exploration dataset, it was found that the best combination of sensors was obtained by combining EDA and Actigraphy data (with the TSD-BCV feature set), each example has a shape varying between $11\times 240$ and $11\times 288$ (Feature $\times$ Time).



\begin{figure*}[!htbp]
\centering
\includegraphics[width=\linewidth]{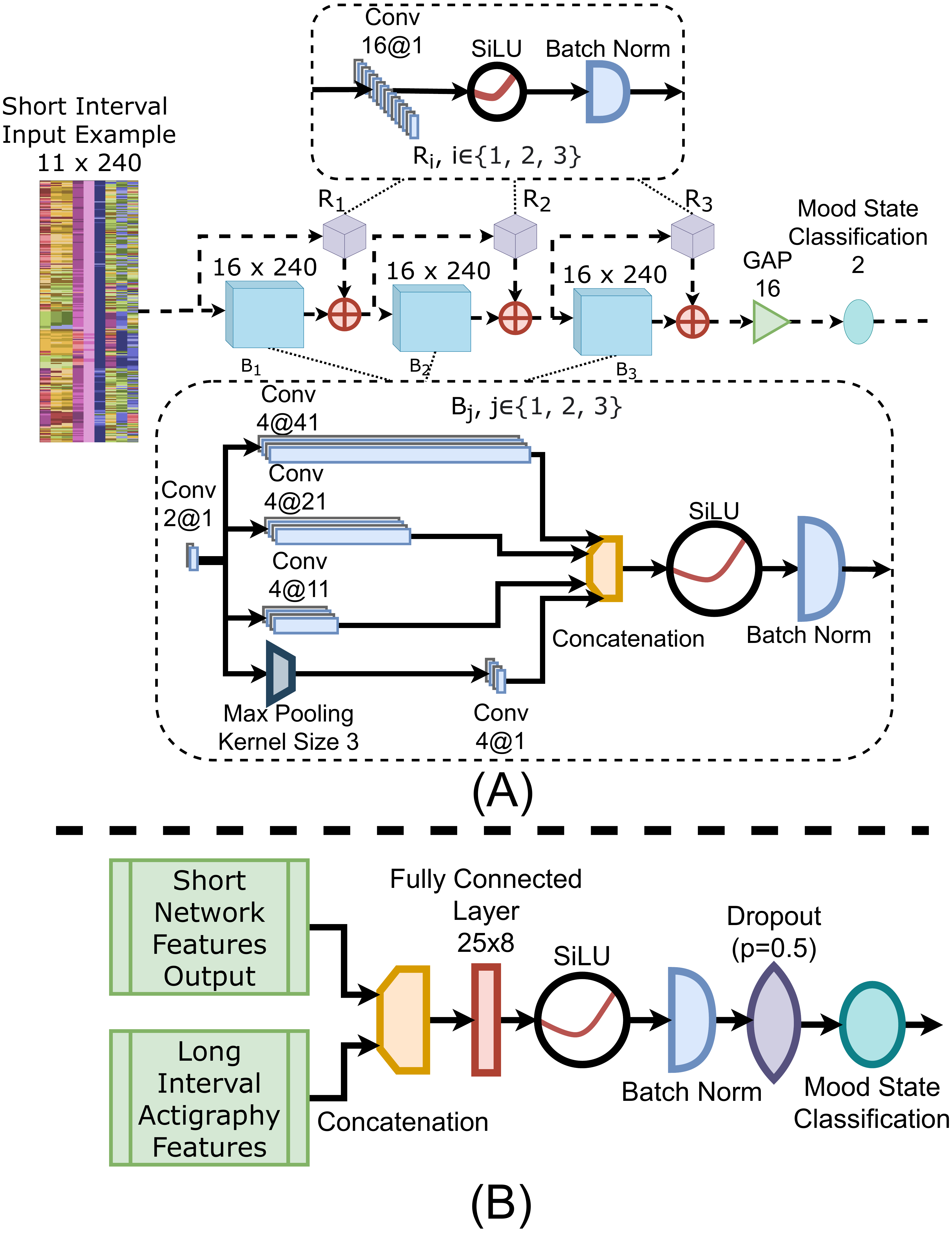}
\caption{(A) The network's architecture employed for Mood-State Bipolar classification using short intervals containing 2830 learnable parameters. In this figure, $R_i$ refers to the ith residual block ($i\in\{1, 2, 3\}$), while $B_j$ refers to the jth InceptionTime block ($j\in\{1, 2, 3\}$). Conv refers to a convolutional layer and GAP refers to the Global Average Pooling operation. Finally, the plus signs refer to an element-wise summation. (B) Short-Long Network's architecture using 5689 parameters. The features from the Short Network corresponds to the output of the global average pooling operation in the Short Network.
} 
\label{fig:CombinedNetwork}
\end{figure*}

Figure~\ref{fig:CombinedNetwork}-(A) details the proposed network's architecture which is refered to as the \textit{Short Network}. RangerLars~\cite{RangerLars} is employed for the network's optimization with a batch size of 128. The learning rate (lr=0.0037) was selected from the exploration dataset by random search using a uniform random distribution on a logarithmic scale between $10^{-6}$ and $1$ with 50 candidates (each candidate was evaluated five times). Mini-batches are built using a bucket approach where sequences of the same length are grouped together. Early stopping, with a patience of 20 epochs is applied by using 10\% of the participants in the training set as a validation set (randomly selected). Additionally, learning rate annealing, with a factor of five and a patience of ten was also used. 


\subsection{Short-Long Interval Classification}
Features extracted from biosignals spanning different time intervals represent different characteristics of human behavior~\cite{petter_actigraphy_ml_depression_classification, shaffer_hrv_features}. Therefore, this work proposes leveraging features extracted from both short (five minutes) and long (20h) periods. To do so, first a Short Network is trained as described in the previous section. After training, the network's weights are frozen and a second network is created, which is shown in Figure~\ref{fig:CombinedNetwork}-B. This network takes the concatenation of the long-interval features and the learned features from the Short Network (directly after the Global Average Pooling) as input and will be referred thereafter as the \textit{Short-Long Network}. The Short-Long Network's architecture was built using the exploration dataset and training procedure is as described in the previous section. Note however, that this time, the interval length is static (20h).


\subsection{Ensemble Method}
As mentioned in~\cite{time_series_inception_2020}, InceptionTime networks can exhibit high variance in terms of performance between training
and therefore can benefit from an ensemble method approach. Consequently, this work also considers an ensemble of five networks for mood-state classification for both the Short and the Short-Long network. The predicted state will thus be the average prediction over the five networks' output. These methods will be referred to as the \textit{Short Ensemble Networks} and the \textit{Short-Long Ensemble Networks} respectively. 

It should be noted that ensemble approaches substantially augment both training and inference time of the model. However, in the current context, mood-states evolve over a period orders of magnitude higher than the latency added by considering ensemble methods (which take less than a second for inference). 
Consequently, the considered ensemble approaches do not reduce the practical application of the proposed method within this work's context.


The methods to extract the feature sets and networks implementation are \href{https://github.com/UlysseCoteAllard/LongShortNetworkBipolar}{available here.}

\section{Experiments and Results}
\label{results_section}
In this paper, accuracy represents the per-participant mean percentage of correctly classified classes averaged over all participants (i.e. each participant's contribution to the accuracy score is weighted equally regardless of the number of examples provided by said participant). Note that, given the slight class imbalance on a per-participant basis of the considered dataset, a classifier systematically predicting the most common class would achieve an accuracy of 53.19\%.

\subsection{Long Interval}
\begin{figure*}[!htbp]
\centering
\includegraphics[width=\linewidth, trim={0 6.9cm 0 0},clip]{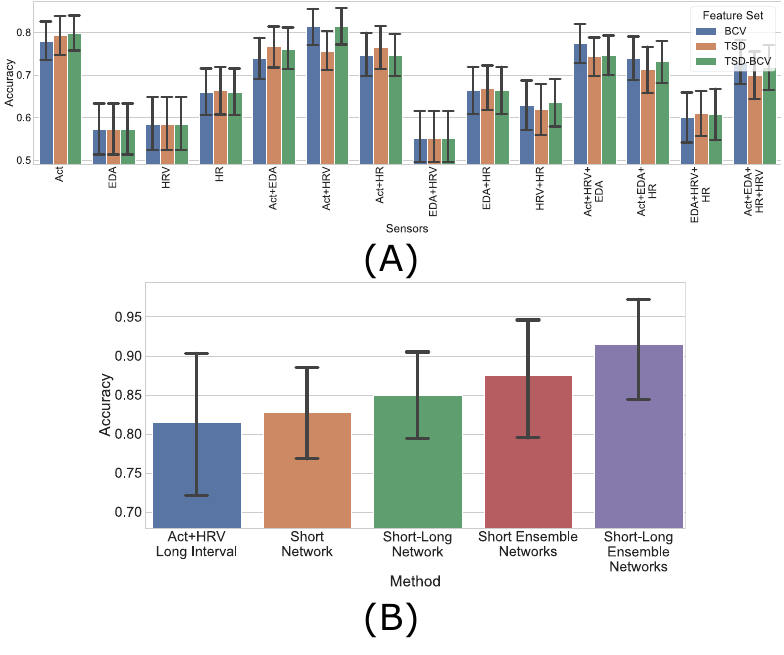}
\caption{
All results are given as an average of 20 runs. Act, EDA, HR and HRV stands for: Actigraphy, Electrodermal Activity, Heart Rate and Heart Rate Variability respectively. BCV, TSD and TSD-BCV correspond to the three feature set considered in this work. (A) - Average accuracy over the 47 participants of the best feature set and classifier combination for each sensor and sensor combinations. (B) - Comparisons between the best performing Long Interval method (Act and HRV from the TSD-BCV feature set using the Linear-SVM classifier) and the proposed Short Interval Methods.
}
\label{fig:results}
\end{figure*}

Figure~\ref{fig:results}-A presents a comparison of the accuracy for mood-state recognition from the different modalities available on the E4 (and combinations of these modalities). For the sake of concision, only the best performing classifier and feature set for each sensor (and their combination) is reported (extended results are provided in Appendix-B).



\subsection{Short and Short-Long Interval}
Figure~\ref{fig:results}-B shows a bar graph comparing the best performing Long Interval method (Actigraphy with TSD-BCV using the Linear-SVM classifier) against the Short Network, Short-Long Network and their ensemble variants.

Following~\cite{statistical_test_classification_demsar_updated}, a two-step statistical procedure using Friedman followed by Finner's post-hoc test was applied. 
First, Friedman's test ranks the algorithms against each other. Then, Finner's post-hoc test is applied (n=47), using the best ranked method as the control method. Finner's null hypothesis is that the mean of the results of the control method against the other methods is equal (compared in pairs).
The null hypothesis is rejected when $p<0.05$. 
Overall, the Short-Long Ensemble Networks obtained the highest average accuracy at 91.59\%$\pm$22.02\% and was the best ranked. Further, the difference between the Short-Long Ensemble Networks and the Actigraphy Long Interval, Short and Short-Long Network was statistically significant ($p=0.01754$, $p<0.00001$ and $p=0.00001$ respectively). No statistically significant difference was found between the Short Ensemble Networks and the Short-Long Ensemble Networks. Appendix-B details these results in a table format.


\section{Discussion}
\label{discussion_section}

Bipolar disorder is a heterogeneous diagnosis~\cite{carvalho2020bipolar}. Consequently, although there are certain common diagnostic criteria, the disorder can manifest widely differently across humans resulting in large behavioral variations during a manic episode. This behavioral variability makes the task of mood-state recognition inherently challenging. Thus, the capability of automatically detecting mood-states in people with bipolar disorders in an objective and non-intrusive way would vastly improve patient outcomes.
This paper proposes leveraging wrist-worn sensors in an effort to meet the challenge. From a clinical perspective, contrastingly to the current cross-sectional mood assessment methods, such an approach could reduce the resource burden and provide evaluations over longer time periods, thereby providing a more comprehensive view of the patients' mood-state. 

For the Long Interval, 336 model combinations were tested (14 possible sensor combinations $\times$ 3 feature sets $\times$ 8 classifiers). Consequently, one should expect that some form of indirect overfitting took place. This was however necessary to get an overall and meaningful picture of the interaction of the different modalities with each other. Additionally, these experiments enable this work to provide a competitive comparison basis of more traditional approaches against the proposed Short and Short-Long networks and their ensemble variants.

The long interval approach was not able to effectively leverage the EDA and HRV features when considered alone. 
In contrast, when combining the features extracted from the actigraphy and HRV, an average accuracy of 81.54\% over 47 participants can be achieved.
Similarly, from the results obtained on the exploration dataset, the combination of Actigraphy+EDA was essential in achieving the best performance. When testing a version of the Short Network using only the Actigraphy data, the performance degraded to around 79\% compared to $\sim$83\% with the proposed EDA+Actigraphy scheme. These results indicate the advantage of considering a multi-sensor approach for mood-state classification, although more work is needed to establish how to best characterize these different signals in a complimentary way.

Overall, using a majority vote over a period of 24h, the proposed Short-Long Ensemble Network was able to correctly classify 45 out of the 47 considered participants (misdiagnosing two manic patients as euthymic). Thus, this work showcases the feasibility of user-independent euthymic-manic state detection in bipolar disorder. 


The long-term goal of this research is to enable automatic early detection of new manic episodes in bipolar individuals at home, through non-intrusive monitoring. Such an instrument would significantly reduce the disease burden by permitting patients or clinicians to implement intervening measures in time to prevent a time-stealing major mood episode. This would revolutionize the monitoring and treatment of bipolar disorder. Therefore, by showing that manic and euthymic episodes can be accurately and automatically distinguished based on biosignals recorded from a smart wristband, this work can be viewed as an important step in this direction.

\subsection{Limitations}
Within this work, the data available for any given participant only spanned $\sim$1 day. This explains the high standard deviation experienced by all methods as the classifiers had, in essence, a single attempt at classifying the participant's state due to the highly correlated data collected for a given recording session. 

As a direct consequence of the absence of intra-subject recording of both states, another limitation of this work is that the training dataset could not provide the learning algorithms with explicit differentiable information between individual variability and mood-state variability. 
An additional distinction of the data considered in this work compared to real-life scenario
in which these models would be applied
, is that all the manic recordings were performed within a clinical environment. Consequently, the participants were receiving active treatment during recording and a certain form of schedule was externally enforced on them, necessarily affecting their behavior. Additionally, the significant increased use of antipsychotic medications (see Table~\ref{table:demographic_characteristic}) in the manic group probably inhibited the elevated energy level commonly associated with mania~\cite{Petter_complexity_variability_2021}. Therefore, the data employed in this work had to contend with an artificially altered gap between manic and euthymic states. 
The impact of which cannot be investigated until these methods are tested in real-life. 
Another limitation related to the recording of manic patients in a clinical environment is that the period of transition between the euthymic and manic state is not currently considered. As such transition periods might hold information that could enable the detection of a future manic episode before its manifestation. Future work will focus on recording an out-of-hospital dataset to test the proposed method in detecting the transition itself instead of the manic episode. 

\section{CONCLUSION}
\label{conclusion_section}
This paper explores bipolar manic-euthymic state recognition using data collected from a wrist-worn sensors. A new feature set for this task was proposed in the TSD-BCV which borrows from both the affective disorder state recognition and the myoelectric-based hand gesture recognition literature. Leveraging actigraphy and HRV data in conjunction with the TSD-BCV, a L-SVM classifier was able to achieve an average accuracy of 81.54\%$\pm$32.39\% over 47 participants (22 manic and 25 euthymic). Further, a new ensemble method comprised of Short-Long Networks was able to achieve an average accuracy of 91.59\%$\pm$22.02\% on the same dataset by leveraging actigraphy and electrodermal activity data. Thus, showcasing the advantage of a multi-sensor approach for bipolar state-recognition. As current diagnostic practices can be inaccurate and require expert involvement~\cite{bauer2018areas}, our results in automatically predicting mood-state in an unknown patient based on wristband data represent a meaningful step in the development of an instrument to facilitate early detection and intervention of manic episodes.


\section{ACKNOWLEDGMENT}
This work was partially supported by the Research Council of Norway as a part of the INTROMAT project (grant agreement 259293)

\bibliographystyle{IEEEtran}
\bibliography{main}

\vspace{-0.28cm}
\section{Authors}
\begin{IEEEbiography}{Ulysse Côté-Allard}{\,} is a Researcher at the University of Oslo, Norway. His main research interests include rehabilitation engineering, biosignal-based control, and human-robot interaction. Contact him at ulysseca@uio.no.
\end{IEEEbiography}

\begin{IEEEbiography}{Petter Jakobsen}{\,} is a PhD candidate at the University of Bergen, Norway. His main research interests include e-health and the characterization of bipolar disorder from actigraphy. Contact him at petter.jakobsen@helse-bergen.no.
\end{IEEEbiography}


\begin{IEEEbiography}{Andrea Stautland}{\,} is a medical doctor and PhD student at the University of Bergen, Norway. She has researched bipolar disorder since 2017 and has a special interest in biomarkers and personalized medicine. Contact her at andrea.stautland@uib.no.
\end{IEEEbiography}

\begin{IEEEbiography}{Tine Nordgreen}{\,} is an associate professor at the University of Bergen, Norway and the leader of the INTROMAT project. Her research centers around developing evidence-based Internet-based treatment within mental health. Contact her at tine.nordgreen@uib.no. 
\end{IEEEbiography}

\begin{IEEEbiography}{Ole Bernt Fasmer}{\,} is a professor at the University of Bergen, Norway. His research interests includes mathematical analyses of time series and the biological aspects of bipolar disorder and ADHD. Contact him at Ole.Fasmer@uib.no.
\end{IEEEbiography}

\begin{IEEEbiography}{Ketil Joachim Oedegaard}{\,} is a professor at the University of Bergen, Norway and leads the Affective Research Group in Bergen. His research centers around affective disorders and genetics. Contact him at Ketil.Odegaard@uib.no. 
\end{IEEEbiography}

\begin{IEEEbiography}{Jim Tørresen}{\,} is a professor at the University of Oslo, Norway. His research interests include machine learning, bio-inspired computing and robotics. Contact him at jimtoer@ifi.uio.no.
\end{IEEEbiography}

\section{Appendices}
\subsection{Appendix A}
The hyperparameters considered for each classifier for the long interval case were as follows: 

\begin{itemize}
\item KNN: The number of possible neighbors considered were 1, 3, 5, 11, 21. The metric distances considered were the Manhattan distance, the Euclidean distance and the Minkowski distance of the third degree. 

\item DT: The quality of the split was measured either by the Gini Impurity or its entropy. The maximum number of features considered were both the square root and the $log_2$ of the total number of feature fed to the decision tree. The tree could either have a maximum depth of 1, 2, 3, 5, 10 or an infinite maximum depth. Finally, the minimum sample split was taken from a uniform distribution between 0 and 1. 

\item RF: The range of the number of trees considered were 10, 50, 100, 500 or 1000. The other considered hyperparameters were the same as for the DT classifier. 

\item AdaBoost: The number of estimators were one of 1, 10, 50, 100, 200. The learning rate was drawn from a logarithm uniform distribution between $10^-3$ and $10^0$. 

\item SVM: For both the linear and RBF kernel, the soft margin tolerance (C) was chosen between $10^-4$ and $10^3$ on a logarithm uniform distribution. Additionally, for the RBF kernel, the $\gamma$ hyperparameter was also selected on a logarithm uniform distribution between $10^{-4}$ and $10^3$.
\end{itemize}

\subsection{Appendix B}
\subsubsection{Results Long Interval}
The best classifier obtained for every combination of sensors and feature set considered for the long interval are given in Table~\ref{table:long_results}.
\begin{table*}[!ht]
\centering
\caption{Average accuracy over the 47 participants of the best classifier combination for every feature set, sensors and their combinations.
For each participant the average accuracy over 20 runs is given.}
\begin{tabular}{ccccccc}
\hline
Modalities & Feature Set & \begin{tabular}[c]{@{}c@{}}Best\\ Classifier\end{tabular} & Accuracy & SD & Friedman's Rank & H0 (p-value) \\ \hline
\multirow{3}{*}{Actigraphy} & BCV & L-SVM & 78.02\% & 35.73\% & 2.06 & 1 \\
 & TSD & L-SVM & 79.23\% & 36.45\% & 2.02 & 1 \\
 & \textbf{TSD-BCV} & \textbf{L-SVM} & \textbf{79.80\%} & \textbf{33.71\%} & \textbf{1.92} & \textbf{-} \\ \cline{2-7} 
\multirow{3}{*}{EDA} & BCV & AdaBoost & 54.13\% & 46.57\% & 2.02 & 1 \\
 & TSD & RBF-SVM & 57.16\% & 45.76\% & 2.05 & 1 \\
 & \textbf{TSD-BCV} & \textbf{AdaBoost} & \textbf{56.81\%} & \textbf{46.29\%} & \textbf{1.93} & - \\ \cline{2-7} 
\multirow{3}{*}{HRV} & BCV & KNN & 57.15\% & 38.48\% & 2.02 & N\textbackslash{}A \\
 & \textbf{TSD} & \textbf{AdaBoost} & \textbf{58.45\%} & \textbf{47.27\%} & \textbf{1.94} & - \\
 & TSD-BCV & AdaBoost & 57.94\% & 47.12\% & 2.04 & N\textbackslash{}A \\ \cline{2-7} 
\multirow{3}{*}{HR} & \textbf{BCV} & \textbf{LDA} & \textbf{66.00\%} & \textbf{43.14\%} & \textbf{1.94} & - \\
 & TSD & L-SVM & 66.38\% & 45.28\% & 2.13 & 1 \\
 & \textbf{TSD-BCV} & \textbf{LDA} & \textbf{66.00\%} & \textbf{43.14\%} & \textbf{1.94} & - \\ \cline{2-7} 
\multirow{3}{*}{Act. + EDA} & BCV & L-SVM & 73.86\% & 36.94\% & 2.13 & 1 \\
 & TSD & L-SVM & 76.73\% & 38.18\% & 1.96 & 1 \\
 & \textbf{TSD-BCV} & \textbf{L-SVM} & \textbf{76.06\%} & \textbf{36.98\%} & \textbf{1.92} & - \\ \cline{2-7} 
\multirow{3}{*}{Act. + HRV} & \textbf{BCV} & \textbf{L-SVM} & \textbf{81.41\%} & \textbf{32.20\%} & \textbf{1.95} & - \\
 & TSD & QDA & 75.69\% & 36.60\% & 1.97 & 1 \\
 & TSD-BCV & L-SVM & 81.54\% & 32.39\% & 2.09 & 1 \\ \cline{2-7} 
\multirow{3}{*}{Act. + HR} & BCV & L-SVM & 74.72\% & 38.55\% & 2.03 & 1 \\
 & TSD & L-SVM & 76.49\% & 38.71\% & 1.99 & 1 \\
 & \textbf{TSD-BCV} & \textbf{LDA} & \textbf{74.56\%} & \textbf{39.74\%} & \textbf{1.98} & - \\ \cline{2-7} 
\multirow{3}{*}{EDA + HRV} & BCV & AdaBoost & 55.80\% & 46.58\% & 1.99 & 1 \\
 & \textbf{TSD} & \textbf{AdaBoost} & \textbf{55.22\%} & \textbf{46.59\%} & \textbf{1.90} & - \\
 & TSD-BCV & KNN & 54.60\% & 43.40\% & 2.11 & 1 \\ \cline{2-7} 
\multirow{3}{*}{EDA + HR} & \textbf{BCV} & \textbf{LDA} & \textbf{66.54\%} & \textbf{44.59\%} & \textbf{1.96} & - \\
 & TSD & KNN & 66.91\% & 41.22\% & 2.09 & 1 \\
 & \textbf{TSD-BCV} & \textbf{LDA} & \textbf{66.54\%} & \textbf{44.59\%} & \textbf{1.96} & - \\ \cline{2-7} 
\multirow{3}{*}{HRV + HR} & BCV & RBF-SVM & 63.02\% & 44.53\% & 2.00 & 1 \\
 & \textbf{TSD} & \textbf{L-SVM} & \textbf{62.08\%} & \textbf{45.86\%} & \textbf{1.92} & \textbf{-} \\
 & TSD-BCV & KNN & 63.69\% & 41.99\% & 2.08 & 1 \\ \cline{2-7} 
\multirow{3}{*}{\begin{tabular}[c]{@{}c@{}}Act. + EDA + \\ HRV\end{tabular}} & BCV & L-SVM & 77.45\% & 34.14\% & 2.00 & 1 \\
 & \textbf{TSD} & \textbf{QDA} & \textbf{74.32\%} & \textbf{36.04\%} & \textbf{1.90} & - \\
 & TSD-BCV & L-SVM & 74.66\% & 36.15\% & 2.10 & 1 \\ \cline{2-7} 
\multirow{3}{*}{\begin{tabular}[c]{@{}c@{}}Act. + EDA +\\ HR\end{tabular}} & \textbf{BCV} & \textbf{RBF-SVM} & \textbf{73.92\%} & \textbf{40.96\%} & \textbf{1.93} & - \\
 & TSD & L-SVM & 71.32\% & 41.77\% & 2.00 & 1 \\
 & TSD-BCV & AdaBoost & 73.14\% & 39.84\% & 2.07 & 1 \\ \cline{2-7} 
\multirow{3}{*}{\begin{tabular}[c]{@{}c@{}}EDA + HRV +\\ HR\end{tabular}} & BCV & QDA & 59.98\% & 45.95\% & 1.98 & 1 \\
 & \textbf{TSD} & \textbf{KNN} & \textbf{60.96\%} & \textbf{43.17\%} & \textbf{1.94} & - \\
 & TSD-BCV & RBF-SVM & 60.83\% & 46.78\% & 2.09 & 1 \\ \cline{2-7} 
\multirow{3}{*}{\begin{tabular}[c]{@{}c@{}}Act. + EDA +\\ HRV + HR\end{tabular}} & \textbf{BCV} & \textbf{L-SVM} & \textbf{73.03\%} & \textbf{39.76\%} & \textbf{1.89} & - \\
 & TSD & L-SVM & 69.91\% & 41.97\% & 2.09 & 1 \\
 & TSD-BCV & L-SVM & 71.76\% & 40.46\% & 2.02 & 1 \\ \hline
\end{tabular}
\vspace{0.3cm}
\\Two-step statistical procedure using Friedman's rank test followed by Finner Post-hoc test using the best ranked method as comparison basis. Null hypothesis rejected when H0=0 (p$<$0.05). 
\label{table:long_results}
\end{table*}

\newpage

\subsubsection{Comparison Short, Long and Short-Long intervals}
Table~\ref{table:comparison_results_networks} shows a comparison between the best performing combination of classifier/feature set/sensors for the long interval and the Short, Short-Long and their ensemble variants. 

\begin{table*}[!htp]
\centering
\caption{Comparison of the best found classification scheme for the long interval with the Short and Short-Long networks and their ensemble variants. For each participant the average accuracy over 20 runs is given.}
\begin{tabular}{@{}cccccc@{}}
\toprule
 & \begin{tabular}[c]{@{}c@{}}Best Overall\\ Long Interval\\ (Actigraphy + HRV,\\ L-SVM,\\ TSD-BCV)\end{tabular} & \begin{tabular}[c]{@{}c@{}}Short \\ Network\end{tabular} & \begin{tabular}[c]{@{}c@{}}Short-Long\\ Network\end{tabular} & \begin{tabular}[c]{@{}c@{}}Short Ensemble\\ Network\end{tabular} & \begin{tabular}[c]{@{}c@{}}Short-Long\\ Ensemble\\ Networks\end{tabular} \\ \midrule
Accuracy & 81.54\% & 82.80\% & 84.89\% & 87.45\% & \textbf{91.59\%} \\
Standard Deviation & 31.53\% & 20.92\% & 18.69\% & 27.44\% & \textbf{22.02\%} \\
Friedman's Rank & 2.93 & 3.83 & 3.63 & 2.50 & \textbf{2.12} \\
H0 (p-value) & 0 (0.01754) & 0 (\textless{}0.00001) & 0 (0.00001) & 1 & \textbf{-} \\
Cohen's Dz & 0.26 & 0.71 & 0.64 & 0.27 & \textbf{-} \\ \bottomrule
\end{tabular}
\vspace{0.3cm}
\\Two-step statistical procedure using Friedman's rank test followed by Finner Post-hoc test using the best ranked method as comparison basis. Null hypothesis rejected when H0=0 (p$<$0.05).
\label{table:comparison_results_networks}
\end{table*}

\end{document}